\pdfoutput=1
\documentclass{article} 
\usepackage{nips15submit_e,times}
\usepackage{hyperref}
\usepackage{url}
\usepackage{amsmath}
\usepackage{amssymb}
\usepackage{array,multirow}
\usepackage{todonotes}
\usepackage{graphicx}
\usepackage{latexsym}
\usepackage{xcolor,colortbl}
\usepackage[acronym]{glossaries}
\usepackage{float}
\restylefloat{table}

\def\None{\ensuremath\text{\it None}}

\newacronym{LSTM}{LSTM}{Long Short Term Memory}
\newacronym{SLU}{SLU}{Spoken Language Understanding}
\newacronym{ASR}{ASR}{Automatic Speech Recognition}
\newacronym{DST}{DST}{Dialog State Tracker}
\newacronym{DSTC}{DSTC}{Dialog State Tracking Challenge}
\newacronym{HIS}{HIS}{Hidden Information State}
\newacronym{CRF}{CRF}{Conditional Random Field}
\newacronym{ME}{ME}{Maximum Entropy}
\newacronym{SDS}{SDS}{Spoken Dialogue System}
\newacronym{NN}{NN}{Neural Network}

\newcommand{\ignore}[1]{}

\newcommand{\superscript}[1]{\ensuremath{^{\textrm{#1}}}}
\def\sharedaffiliation{\end{tabular}\newline\begin{tabular}{c}}

\def\dg{\superscript{\dag}}
\def\ddg{\superscript{\ddag}}

\title{Hybrid Dialog State Tracker}

\author{
Miroslav Vodol{\'a}n\dg\ddg\\
Charles University in Prague\ddg\\
Faculty of Mathematics and Physics\\
Malostranske nam. 25, 11800, Prague\\
\And
Rudolf Kadlec\dg and Jan Kleindienst\dg\\
IBM Watson\dg\\
V Parku 4\\
Prague 4, Czech Republic\\
 \sharedaffiliation
  \begin{tabular}{ccc}
    \texttt{\{mvodolan, rudolf\_kadlec, jankle\}@cz.ibm.com} \\
    \texttt{vodolan@ufal.mff.cuni.cz}
  \end{tabular}
}

\nipsfinalcopy 

\begin{document}

\maketitle

\begin{abstract}
This paper presents a hybrid dialog state tracker that combines a rule based and a machine learning based approach to belief state tracking. Therefore, we call it a hybrid tracker. The machine learning in our tracker is realized by a \gls{LSTM} network. To our knowledge, our hybrid tracker sets a new state-of-the-art result for the \gls{DSTC} 2 dataset when the system uses only live SLU as its input. 
\end{abstract}

\section{Introduction}

\Glspl{SDS} consist of many modules, one of which is a \gls{DST}. \Gls{DST} is responsible for accumulating evidence throughout the dialogue and estimating current true user's goal. The user goal estimate is subsequently used by other modules of the \gls{SDS}, e.g., by a policy module that picks the next best action.

Recently proposed \glspl{DSTC}~\cite{williams-EtAl:2013:SIGDIAL,henderson-thomson-williams:2014:W14-43,DSTC3}
provide a shared testbed with datasets and tools for evaluating of dialog
state tracking methods. It abstracts away the subsystems of end-to-end
spoken dialog systems, focusing only on the dialog state tracking.
It does so by providing datasets of ASR and SLU outputs on slot-filling tasks with reference transcriptions, together with annotation on the level of dialog acts and user goals.

The last three dialog state tracking challenges~\cite{williams-EtAl:2013:SIGDIAL,henderson-thomson-williams:2014:W14-43,DSTC3} were dominated by machine learning based trackers~\cite{Lee2013,Williams2014,Henderson2014b}. However, when we consider the case where all trackers have the same \gls{SLU} input, some rule based trackers~\cite{wang-lemon:2013:SIGDIAL,kadlec2014ibm,sun-EtAl:2014:W14-43,kadlec2014knowledge} achieved performance comparable to the top trackers. In this work, we aim to unite the best of the both worlds --- high accuracy of the machine learning trackers and better interpretability of the rule based trackers. A similar research direction was recently explored in~\cite{Sun2015}.
The core of our proposed tracker consists of several update rules that use a few parameters that are computed by a recurrent neural network. We show that on the DSTC2 dataset our hybrid tracker achieves the state-of-the-art performance among the systems that use the original live \gls{SLU}. Note that the \glspl{DST} that also use \gls{ASR} output as additional feature achieve even better tracking accuracy. We will add these features in a future work.

For evaluation of the tracker we chose the \gls{DSTC}2 because it contains complex dialogs with changes of the user's goal and it also provides a lot of training data. Dialogs in the \gls{DSTC}1 did not have frequent user goal changes and the \gls{DSTC}3 had only a limited training dataset. The challenges also differ in their domains. The \gls{DSTC}1 dataset is collected from system providing bus routes, the \gls{DSTC}2 is focused on restaurant domain and the \gls{DSTC}3 combines restaurant and hotel domains.

In the next section we describe the architecture of our Hybrid tracker. Then we evaluate the tracker on the DSTC2 dataset and conclude the paper with an outline of our future work.


\section{Hybrid dialog state tracker model}
In our previous work~\cite{kadlec2014knowledge}, we introduced a belief tracker based on a few simple rules which scored second in the joint slot accuracy in \gls{DSTC}3 and its slightly modified version has the state-of-the-art accuracy on this dataset. Here we simplify the original rules for per slot tracking (in contrast with~\cite{kadlec2014knowledge} where we tracked the slots jointly) and we add the machine learning component that provides parameters for these rules. We call the resulting architecture a hybrid tracker.

The tracker operates on a probability distribution over values for each slot separately. For each turn, the tracker generates these distributions reflecting the user's goals based on the last machine action, the observed user actions, the probability distributions in the previous turn and the hidden state $l_{t-1}$ of a recurrent network $L$ from the previous turn. The probability distribution $h^s_t$ for a single slot $s$ and turn $t$ is represented by a vector indexed by possible values of the slot $s$. The joint belief state is represented by the probability distribution over Cartesian product for each slot. 

\begin{figure}
    \begin{center}
	    \includegraphics[scale=0.9]{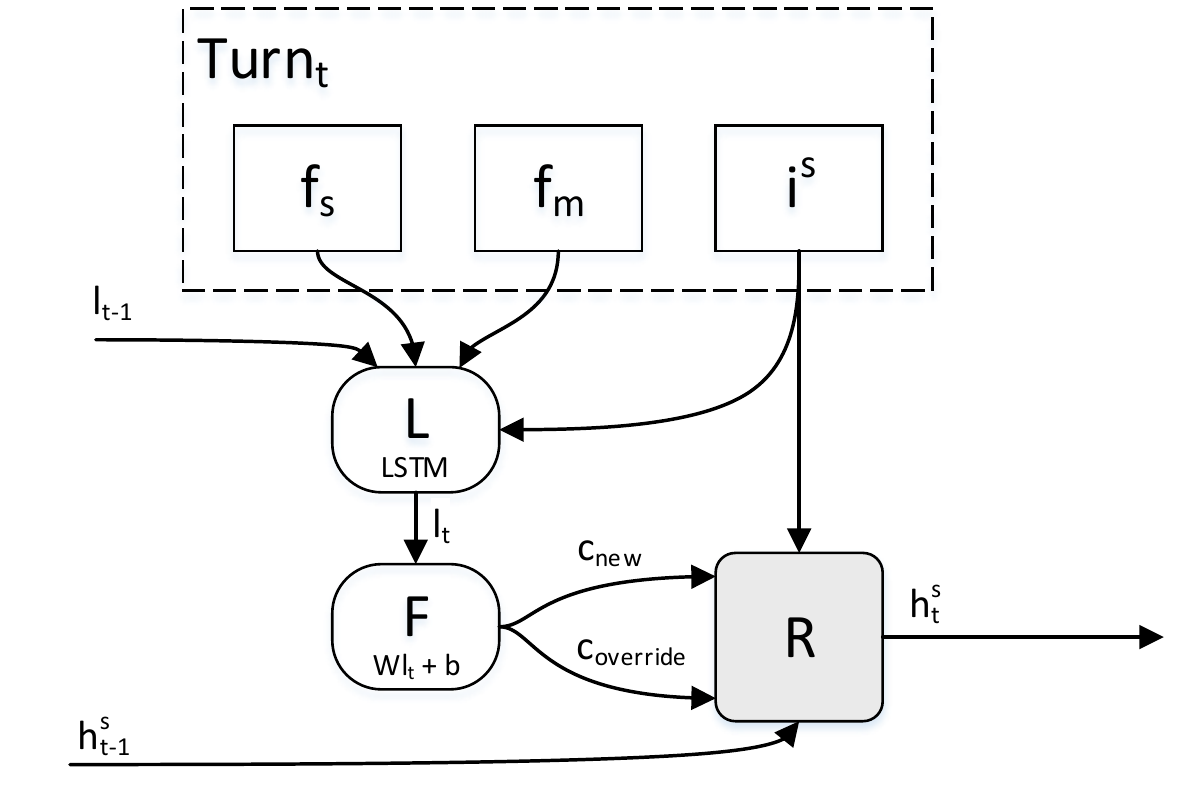}
	\end{center}
	
    \caption{The structure of the Hybrid tracker for the turn $t$. It is a recurrent model which uses probability distribution $h^s_{t-1}$ and $l_{t-1}$ from the previous turn. Inputs of the machine learning part of the model (represented by functions $L$ and $F$) are the features indicating the tracked slot $f_s$ and the features $f_m$ extracted from the machine actions. The features are used to produce values of parameters $c_{\text{new}}$ and $c_{\text{override}}$ for the $R$ function.}
    \label{fig:tracker_structure}
\end{figure}

In the following notation $i^s_t$ denotes a user action pre-processed into a probability distribution of informed values for the slot $s$ and turn $t$. During the pre-processing every \textit{Affirm()} from \gls{SLU} is transformed into \textit{Inform(slot=value)} according to the machine action $m$. Further, we introduce a function corresponding to the simplified rules (fully described in Sec.~\ref{ssec:knowledge_based_part})
$h^s_t=R(h^s_{t-1}, i^s, c_{new}, c_{override})$, which is a function of a probability distribution in the previous turn, the pre-processed user action and two parameters which control how the new probability distribution $h^s_t$ is computed.
The next function $l_t=L(l_{t-1}, f_s, f_m, i^s)$ is recurrent and takes its own output $l_{t-1}$ from the previous turn, the features $f_s$ indicating the tracked slot, the features $f_m$ representing machine actions and the pre-processed user action $i^s$ of the turn $t$. The output of the recurrent network is then linearly transformed by $F(l_t)$ to parameters $c_{\text{new}}$ and $c_{\text{override}}$ for $R$. The structure of the tracker is shown in Figure~\ref{fig:tracker_structure}.


In the next subsection, 
we will describe the rule based component of the Hybrid tracker. Afterwards, in Section~\ref{ssec:machine_learned_part}, we will describe the machine learning part of the tracker.

\subsection{Rule-based part}
\label{ssec:knowledge_based_part}
The rule-based part of the tracker represented by the function $R$ consists of several simple update rules 
parametrized by parameters $c_{\text{new}}$ and $c_{\text{override}}$\footnote{These parameters were modelled by a so called \textit{durability} function in our previous tracker~\cite{kadlec2014knowledge}.}. Each of the parameters controls transition probability in a different way:
\begin{itemize}
    \item $c_{\text{new}}$ --- controls how easy it would be to change the belief from hypothesis \textit{None} to an instantiated slot value,
    \item $c_{\text{override}}$ --- models a goal change, that is, how easily it would be to override current belief with a new observation.
\end{itemize}
In this work we compute these parameters by a neural network.
The rule based part of our tracker is specified by following equations. The first equation specifies belief update rule for probability assigned to slot's value $v_1$: 
\begin{equation}
   \label{eq:transition_rules}
    h^s_t[v_1] = h^s_{t-1}[v_1] - \tilde{h}^s_t[v_1] + i^s_t[v_1] \cdot \sum_{v_2 \neq v_1}{ h^s_{t-1}[v_2] \cdot a_{v_1v_2} } 
\end{equation}
Where $\tilde{h}^s_t[v_1]$ corresponds to amount of probability that will be transferred from $h^s_{t-1}[v_1]$ to other slot values in $h^s_{t}$:
\begin{equation}
   \label{eq:transferred_probability}
   \tilde{h}^s_t[v_1] = h^s_{t-1}[v_1]\cdot \sum_{v_2 \neq v_1}{i^s_t[v_2] \cdot a_{v_2v_1}}
\end{equation}
The $a_{v_1v_2}$ is called transition coefficient between values $v_1$ and $v_2$. It controls amount of probability which is transferred from $h^s_{t-1}[v_2]$ to $h^s_t[v_1]$.
\begin{equation}
   \label{eq:transition_coefficients}
   a_{v_1v_2} = 
   \begin{cases}
        v_1 = \None & c_{\text{new}} \\
        v_1 \neq v_2 & c_{\text{override}} \\
   \end{cases}
\end{equation}

As we can see, the $R$ function is differentiable, therefore the machine learned part, described in the following subsection \ref{ssec:machine_learned_part}, can be trained by gradient descent methods together with the rule-based part.

We can find similar update equations in other rule-based trackers, e.g.,~\cite{wang-lemon:2013:SIGDIAL,zilka-EtAl:2013:SIGDIAL,sun-EtAl:2014:W14-43,kadlec2014knowledge}.

\subsection{Machine learned part}
\label{ssec:machine_learned_part}
The machine learning part of our tracker is realized by a \gls{LSTM}~\cite{Hochreiter1997} network. We use recurrent network for $L$ since it can learn to output different values of $c$ parameters for different parts of the dialog (e.g., it is more likely that new hypothesis will arise at the beginning of a dialog). This way, the recurrent network influences the rule-based component of the tracker. Since there are only two parameters that are used by the rule-based part, the tracker's decisions can be easily introspected.

The function $L$ uses the feature $f_s$, which is one-hot representation of the tracked slot and the feature $f_m$ which is a bag of words representation of machine actions. The last feature of the $L$ function is pre-processed user action $i^s$ representing marginal probabilities of informed values for slot $s$.

In our tracker we use one machine learned model that is shared for all slots. However, the model can distinguish between the slots according to $f_s$ feature. The other systems use a different setup where a shared model is trained for all slots and then it is fine-tuned for each separate slot~\cite{henderson-thomson-young:2014:W14-43}.


\section{Evaluation}

\subsection{Method}

 The parameters of the hybrid tracker were trained by SGD with AdaGrad~\cite{duchi2011adaptive} and Adam~\cite{kingma2014adam} weight update rules. This is possible since of all parts of the model are differentiable (including the $R$ function). 
 
 We trained two groups of trackers with different settings. The first group was trained by the AdaGrad algorithm with the learning rate $0.5$ and the gradient clipping with threshold $10$. With this setting the training algorithm produced trackers heavily influenced by random initialization, which is good for later ensembling of the trackers. For the second group we used the Adam update rule with the learning rate $0.01$, $\beta_1$ $0.9$ and $\beta_2$ $0.999$. These settings are much more invariant to random initialization therefore we randomly masked $f_m$ features to get set of different trackers. Both groups used $L$ function with $5$ \gls{LSTM} cells and $tanh$ as the activation function.

 From each dialog in the \textit{dstc2\_train} data ($1612$ dialogs) we extracted training samples for the slots \textit{food}, \textit{pricerange} and \textit{area} and used all of them to train each tracker. The training data was also used for selection of $f_m$ features. We selected only those words from machine action\footnote{The machine action is represented by dialog acts.} that appeared more than $5$ times. This gives us the total number of $421$ $f_m$ features and $3$ $f_s$ features (one per \textit{food}, \textit{pricerange} and \textit{area} slot). 
 
 The evaluated model was an ensemble of multiple trackers that were combined by averaging. Similar approach proved to be useful also in other RNN based trackers~\cite{henderson-thomson-young:2014:W14-43,Zilka2015}. For the ensemble, we used $100$ trackers randomly selected from both tracker groups containing $115 + 143$ trackers. 
 
 We evaluated $10$ different ensembles and selected the one with the best performance on validation \textit{dstc2\_dev} ($506$ dialogs) data, which is reported in subsection~\ref{ssec:results}. Our tracker did not track the \textit{name} slot because it hurts validation performance. Therefore, we always set value for the \textit{name} slot \textit{None}. The mean accuracy of the $10$ ensembles on \textit{dstc2\_test} data ($1117$ dialogs) is $0.7448$ with the standard deviation $0.0006$.
 
The models were implemented using Theano~\cite{Bastien-Theano-2012} and Blocks~\cite{VanMerrienboer2015}. 

\begin{table}[H]
\begin{center}
\bgroup
\def\arraystretch{1.05}
\setlength{\tabcolsep}{3pt}
\begin{tabular}{ll|c||cc|cc||cc}
 & & & \multicolumn{2}{c}{test2}  \\ \cline{3-5} 
 & & ASR & Acc. & L2 & post DSTC\\[0.1cm] \hline
 
& Focus baseline    &           &   \textbf{.719}   &   \textbf{.464}    \\ 
& HWU baseline     &   &   .711  &   .466   \\ \hline
& DSTC2 stacking ensemble \cite{henderson-thomson-williams:2014:W14-43} & $\surd$ & \textbf{.798} & \textbf{.308} & $\surd$\\ \hline
& Williams \cite{williams:2014:W14-43} & $\surd$	&   \textbf{.784}	&   .735	\\
& Henderson et al. \cite{henderson-thomson-young:2014:W14-43} \ignore{t4e3} & $\surd$ & .768	& \textbf{.346} \\
& Yu et al.~\cite{Yu2015}  & $\surd$ & .762	& .436 & $\surd$ \\
& YARBUS \cite{Fix2015} & $\surd$ & .759 & .358 & $\surd$ \\
& Sun et al. \cite{sun-EtAl:2014:W14-43} \ignore{t7e4} & $\surd$   &   .750 &	.416 \\
\hline
\hline
& \textbf{Hybrid Tracker -- This work} & & \textbf{.745} & .433 & $\surd$ \\
& Williams \cite{williams:2014:W14-43} &	&  .739   &   .721	\\
& Henderson et al. \cite{henderson-thomson-young:2014:W14-43} \ignore{t4e3} & 	& .737 &	\textbf{.406} \\
& Our previous tracker~\cite{kadlec2014knowledge} & & .737 & .429 & $\surd$ \\
& Sun et al. \cite{sun-EtAl:2014:W14-43} \ignore{t7e4} &   &   .735   &   .433 \\
& Smith \cite{smith:2014:W14-43} \ignore{t3e0}   & & .729   & .452  \\
& Lee et al. \cite{lee-EtAl:2014:W14-43} & & .726 & .427 \\
& YARBUS \cite{Fix2015} & & .725 & .440 & $\surd$ \\
& Ren et al. \cite{ren-xu-yan:2014:W14-43} \ignore{t6e2}  &    & .718  &   .437  \\

\hline



\end{tabular}
\egroup
\end{center}
\caption{Joint slot tracking results for various systems reported in the literature. The trackers that used ASR have $\surd$ in the corresponding column. The results of systems that did not participate in DSTC2 are marked by $\surd$ in the "post DSTC" column. The first group shows two  baselines provided in the DSTC2. The second group shows results of an ensemble of all trackers submitted to the challenge. This system achieves the best result among the systems that use both the original SLU and ASR. The third group lists individual trackers that use ASR. The fourth group lists systems that use only the live SLU provided in the original dataset. Our hybrid tracker sets new state-of-the-art result in this category. The best results for a given metric and tracker group are in bold.}
\label{tab:results-dstc2}
\end{table}

\subsection{Results}
\label{ssec:results}


Table~\ref{tab:results-dstc2} shows the results of our hybrid tracker and other top performing trackers known from the literature. In the category of trackers that use only the live SLU features our systems sets the new state-of-the-art with accuracy $0.745$ on \textit{dstc2\_test}. The accuracy of the tracker on \textit{dstc2\_dev} is $0.657$ and $0.767$ on \textit{dstc2\_train}.

\subsection{Discussion}
Evaluation on the DSTC2 dataset shows that our hybrid system that extends rule based tracking core with the machine learning component outperforms the previous best tracker~\cite{williams:2014:W14-43} that used the same SLU input. This result is also interesting since our ML component is relatively lightweight (it has only approx. 10k parameters, the hidden state consist of only 5 neurons) and it influences computation of the rules part by only 2 parameters.   



\section{Future Work and Conclusion}
We have presented a belief tracker that combines our previous tracker with machine learning techniques. It performs better than our previous tracker while still being highly interpretable in comparison with pure neural network approaches. 

However, trackers that use ASR as their input achieve even better accuracy. Therefore the next step will be to add ASR features to our machine learning component. This will hopefully allow us to further improve accuracy of our system.

\section*{Acknowledgments} 
This research was partly funded by the Ministry of Education, Youth and Sports of the Czech Republic under the grant agreement LK11221, and core research funding of Charles University in Prague.

\bibliographystyle{ieeetr}  
\bibliography{hybrid_tracker}

\end{document}